\documentclass[runningheads]{llncs}

\usepackage{eccv}

\usepackage{eccvabbrv}

\usepackage{xcolor}

\usepackage{subcaption} 
\usepackage{graphicx}
\usepackage{booktabs}
\usepackage{amssymb}
\newcommand{\cmark}{\checkmark}

\usepackage{siunitx}
\usepackage{bm}

\usepackage[accsupp]{axessibility}  

\usepackage{tikz}
\usetikzlibrary{positioning,arrows.meta,fit}
\newcommand{\myparagraph}[1]{\vspace{1mm}\noindent\textit{#1}}

\usepackage{hyperref}

\usepackage{orcidlink}

\begin{document}

\title{LangLoc: ``Tell Me What You See''} 



\author{
Shaurya Kishore Panwar\inst{1,2}\thanks{Equal contributions.} \and
Roham Zendehdel Nobari\inst{1,2}$^{*}$ \and
Shirley Feng Yi Lau\inst{1,2}$^{*}$ \and
Abu Bakr Rahman Shaik\inst{1,2}$^{*}$\orcidlink{0009-0006-8630-0339} \and
Manuel Günther\inst{2}\orcidlink{0000-0003-1489-7448} \and
Marc Pollefeys\inst{1,3} \and
Daniel Barath\inst{1,4}
}

\authorrunning{S. ~Panwar et al.}


\institute{
ETH Zürich, Switzerland \and
University of Zürich, Switzerland \and
Microsoft \and
HUN-REN SZTAKI, Hungary
}

\maketitle

\begin{abstract}
    We tackle fine-grained indoor localization from natural language: given a free-form description of one's surroundings, estimate the observer's 2D position and heading within a known 3D environment.
    Language queries are lightweight, privacy-preserving, and need no camera -- yet prior work stops at coarse scene retrieval and cannot resolve an intra-scene pose.
    We close this gap with \emph{LangLoc}, a three-stage pipeline that (i)~retrieves the correct scene via a dual-branch GATv2 encoder with CLIP semantic features, surpassing the previous best by 8 percentage points in Top-1 recall; (ii)~estimates position and heading by scoring a dense floor grid through ray-cast object visibility, reaching a median error of 0.95\,m; and (iii)~resolves residual ambiguity through a Bayesian dialog module that asks targeted yes/no questions and updates a pose posterior until the location is pinpointed.
    To support this task we contribute a benchmark of $13{,}000{+}$ pose-indexed natural-language descriptions over $1{,}300{+}$ indoor 3D scans.
    Code and data will be released. {Project page: \url{https://rzninvo.github.io/Lang-Loc/}.}
\end{abstract}

\section{Introduction}
\label{sec:intro}

Knowing where you are is fundamental to almost every location-aware service: indoor navigation, robot assistance, augmented reality, and emergency response all require an accurate pose estimate.

The dominant localization paradigm today is visual: a device captures an image or video stream, uploads it to a server, and receives a pose estimate in return~\cite{sattler2017activesearch,sarlin2019hloc}. While effective, this approach carries significant drawbacks. Image transmission is bandwidth-heavy, especially indoors where frequent queries are needed. More critically, it is \emph{privacy-invasive}: photos of homes, offices, and hospitals inevitably capture sensitive information that users may not wish to share. Finally, capturing a \emph{useful} image is itself non-trivial -- a photo of a blank wall carries little discriminative information, requiring users to know how to frame an informative shot.

Language offers a compelling alternative. Telling a system ``I'm standing in front of a bookshelf, with a blue sofa on my left and a TV across the room'' is natural, fast, and transmits almost no personally identifiable information. A text description is orders of magnitude smaller than an image, requires no camera or special hardware, and mirrors how people naturally communicate their whereabouts to one another in everyday life. This makes language localization natural in camera-prohibited but digitally-twinned settings such as hospitals, labs, and emergency dispatch. Beyond localization, human-to-agent communication also requires grounding free-form verbal goals (e.g., ``go to the bookshelf and find the red book'') into precise 3D poses for robots, drones, and AR assistants.

Despite this appeal, language-based localization remains largely unsolved. Existing methods address only \emph{coarse scene retrieval} -- identifying which room in a database a description refers to~\cite{kolmet2022text2pos,chen2024whereami}. Resolving a precise pose \emph{within} a scene from language is an open problem: many viewpoints within the same room share similar semantics, differing only in subtle geometric or visibility cues that are difficult to capture in plain text.

We present \emph{LangLoc}, the first pipeline for fine-grained indoor localization from natural language. Given a free-form description and a database of 3D scenes, LangLoc first retrieves the correct scene -- surpassing the prior state of the art by 8 percentage points in Top-1 recall -- and then estimates a 2D floor position and heading within it, achieving approximately $1$\,m median position error. When a description is ambiguous, the system enters an interactive dialog: it asks targeted yes/no questions (\eg ``Is there a chair to the left of the table?'') and updates a Bayesian pose posterior until the location is resolved.

\myparagraph{Contributions.}
\begin{itemize}
    \item \textbf{Scene retrieval.} A dual-branch GATv2 encoder with CLIP features that sets a new SOTA, with an 8\,percentage points improvement over prior work~\cite{chen2024whereami}.
    \item \textbf{Fine-grained localization.} A visibility-based floor-grid scoring method that estimates a 2D position and heading direction from language, achieving ${\approx}1$\,m median error.
    \item \textbf{Dialog-based disambiguation.} An interactive Bayesian refinement module that resolves ambiguous descriptions through targeted yes/no questions.
\end{itemize}

\section{Related Work}
\label{sec:related}

\myparagraph{Visual localization.}
Estimating the 6-DoF camera pose in a known environment is a long-standing problem in computer vision.
Structure-based methods build an explicit 3D map and localize by establishing 2D--3D correspondences followed by PnP solving~\cite{sattler2017activesearch,taira2018inloc,sarlin2019hloc}.
Learned local features~\cite{detone2018superpoint} and matching networks~\cite{sarlin2020superglue} have substantially improved robustness. Hierarchical pipelines that first retrieve candidate images then perform local matching~\cite{sarlin2019hloc} define the dominant paradigm.
An alternative line of work regresses pose directly from a single image via a CNN~\cite{kendall2015posenet,brahmbhatt2018mapnet}, trading some accuracy for architectural simplicity.
Scene coordinate regression methods~\cite{brachmann2017dsac,brachmann2018dsacpp,brachmann2023ace} recover much of this accuracy gap by predicting dense 3D coordinates and integrating a differentiable PnP+RANSAC solver, achieving state-of-the-art single-image localization.
Visual place recognition~\cite{arandjelovic2016netvlad,hausler2021patchnetvlad,berton2022cosplace,alibey2023mixvpr} tackles the related retrieval problem of identifying the closest location in a database, analogous to the coarse stage of our pipeline but with an image query.
Recent work has moved toward lighter map representations: PixLoc~\cite{sarlin2021pixloc} refines poses via learned feature-metric alignment, and OrienterNet~\cite{sarlin2023orienternet} localizes against 2D public maps.
All of the above methods require a visual query.
Our work departs from this paradigm by replacing the image with a natural-language description, probing how much localization accuracy is attainable from language alone.

\myparagraph{Language-based and cross-modal localization.}
Text2Pos~\cite{kolmet2022text2pos} first studied text-to-3D localization by learning a joint embedding between free-form descriptions and large-scale outdoor point clouds for coarse cell retrieval.
In the indoor domain, Chen~\etal~\cite{chen2024whereami} frame the problem as scene retrieval over 3D scene graphs: a text query is parsed into a graph and matched against a database of 3DSSG~\cite{3DSSG2020} graphs to identify the correct room.
Their Text2SGM model demonstrates that scene-graph matching from language is feasible, but it stops at coarse scene identification and does not resolve an intra-scene camera pose.
SceneGraphLoc~\cite{miao2024scenegraphloc} performs cross-modal coarse localization by matching image features to 3D scene graphs with a dual-branch GATv2 encoder -- an architecture we adapt for text-to-graph retrieval.
SGAligner~\cite{sarkar2023sgaligner} learns multi-modal scene-graph embeddings that align 3D, image, and text modalities, further demonstrating the utility of scene graphs as a bridge between modalities.
Our work builds on these foundations but extends the pipeline beyond scene retrieval to fine-grained pose estimation within the retrieved scene.

\myparagraph{Dialog-based localization and embodied navigation.}
In embodied settings, agents often interact with humans via dialog to resolve spatial ambiguity.
Vision-and-Dialog Navigation~\cite{thomason2020visiondialog} introduces cooperative dialogs for object-goal navigation. DiaLoc~\cite{zhang2024dialoc} proposes iterative dialog-based localization, where an agent narrows its pose estimate through clarification questions.
The broader VLN literature~\cite{anderson2018r2r,qi2020reverie,hong2021vlnbert} studies instruction-following in photorealistic environments. Recent LLM-based agents~\cite{zhou2024navgpt} bring explicit reasoning to navigation.
These works motivate our dialog-based disambiguation module, which selects targeted questions to reduce uncertainty over candidate poses.

\myparagraph{3D scene graphs and vision-language grounding.}
Scene graphs~\cite{armeni20193d} encode objects, attributes, and spatial relationships in a unified hierarchical structure.
3DSSG~\cite{3DSSG2020} extends this idea to learned prediction of semantic scene graphs from 3D indoor reconstructions, while incremental methods~\cite{wu2021scenegraphfusion} enable on-the-fly graph construction from RGB-D streams.
Open-vocabulary extensions using foundation models~\cite{gu2024conceptgraphs,peng2023openscene,kerr2023lerf} have further broadened the applicability of scene graphs by removing the need for a fixed vocabulary.
On the language side, ScanRefer~\cite{chen2020scanrefer} and ReferIt3D~\cite{achlioptas2020referit_3d} pair referring expressions with 3D bounding boxes. ScanQA~\cite{azuma2022scanqa} and SQA3D~\cite{ma2023sqa3d} extend this to question answering with spatial reasoning.
Pre-trained 3D--language models~\cite{zhu20233dvista,hong2023_3dllm} align point clouds or multi-view features with text at scale.
These resources target grounding or retrieval rather than viewpoint-conditioned pose estimation, leaving a gap in both data and methods for fine-grained language-based localization that our work aims to fill.

\begin{figure*}[t]
\centering
\makebox[\textwidth][c]{\includegraphics[width=1.09\textwidth]{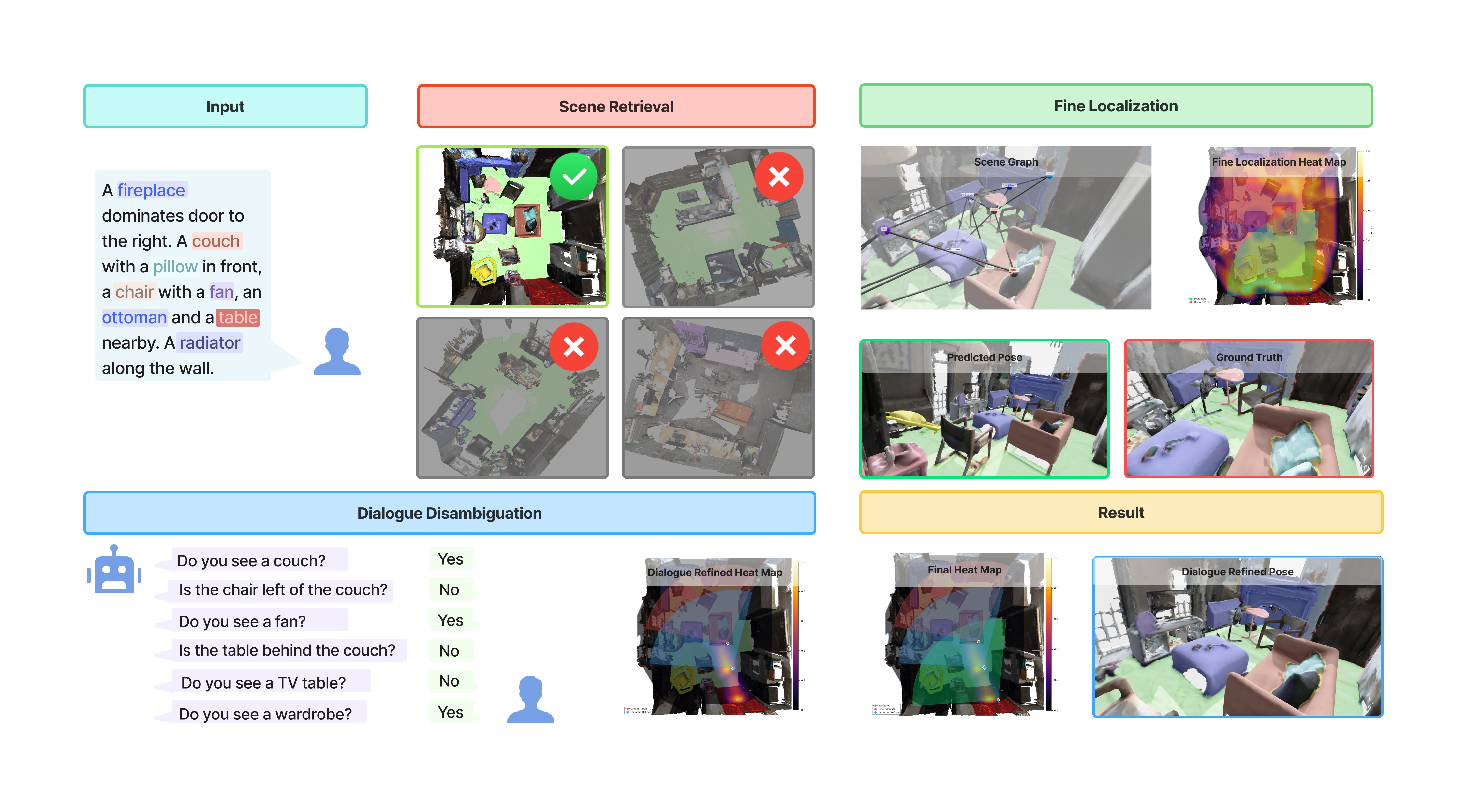}}
\vspace{-10mm}
\caption{\textbf{LangLoc overview.} \emph{Input:} a free-form description mentioning objects such as fireplace, couch, pillow, and fan. \emph{Scene retrieval} (\cref{sec:scene_retrieval} ranks candidate 3D scenes; the correct scene (green, \cmark) is identified from a pool of candidates. \emph{Fine localization} (\cref{sec:fine_localization}) scores a dense floor grid by ray-cast object visibility, producing a heat map over candidate positions; the predicted pose (green frustum) is compared against the ground truth (red frustum). \emph{Dialog disambiguation} (\cref{sec:dialog}) asks targeted yes/no questions (e.g., ``Do you see a couch?'', ``Is the chair left of the couch?'') and iteratively refines the posterior until it concentrates on the correct pose.}
\label{fig:overview}
\end{figure*}

\section{Language-based Localization}
\label{sec:method}

Our pipeline has three stages (\cref{fig:overview}): scene retrieval (\cref{sec:scene_retrieval}), fine localization (\cref{sec:fine_localization}), and dialog-based disambiguation (\cref{sec:dialog}).

\subsection{Problem Formulation}
\label{sec:problem_formulation}

The input is a free-form textual description $T$ of an observer's surroundings; the output is an estimated observer pose within a known indoor environment.
The setting extends language-based scene retrieval from 3D scene graphs~\cite{chen2024whereami}: we not only identify the correct scene but also resolve a precise intra-scene pose.

\myparagraph{3D scene database.}
Let $\mathcal{D}=\{ \mathcal{S}_i \}_{i=1}^{N}$ denote a database of scanned indoor scenes.
Each scene $\mathcal{S}_i$ provides a reconstructed mesh with semantic instance segmentation and a 3D scene graph $G^s_i=(V^s_i,E^s_i)$.
Nodes represent object instances with semantic labels; edges encode typed spatial relations (e.g., \textit{left-of}, \textit{on}).

\myparagraph{Text scene graph.}
Following \cite{chen2024whereami}, we parse $T$ into a text scene graph $G^t=(V^t,E^t)$. Nodes represent objects; edges encode relational statements from the text. This enables structured matching against the database graphs $\{G^s_i\}$.

\myparagraph{Tasks.}
Given $(T,\mathcal{D})$, the system executes three tasks:
\begin{enumerate}
    \item \textbf{Scene retrieval.} Produce a ranked list of candidate scenes $\pi(T) = (i_1,\ldots,i_K)$ by matching $G^t$ against each $G^s_i$.
    \item \textbf{Fine localization.} For the top candidate scene, estimate a 2D floor position $\hat{\mathbf{c}}\in\mathbb{R}^2$ and a unit direction vector $\hat{\theta}\in\mathbb{S}^2$. We denote the full pose as $\hat{\mathbf{p}}=(\hat{\mathbf{c}},\hat{\theta})\in\mathbb{R}^2\times\mathbb{S}^2$ and the ground-truth pose as $\mathbf{p}^\star=(\mathbf{c}^\star,\theta^\star)$.
    \item \textbf{Dialog-based disambiguation.} When the posterior over poses is multi-modal, the system asks targeted yes/no questions about object presence or spatial relations to iteratively reduce ambiguity~\cite{thomason2020visiondialog,zhang2024dialoc}.
\end{enumerate}

\subsection{Language-based Scene Retrieval}
\label{sec:scene_retrieval}

\begin{figure*}[t]
\centering
\makebox[\textwidth][c]{\includegraphics[width=1.09\textwidth]{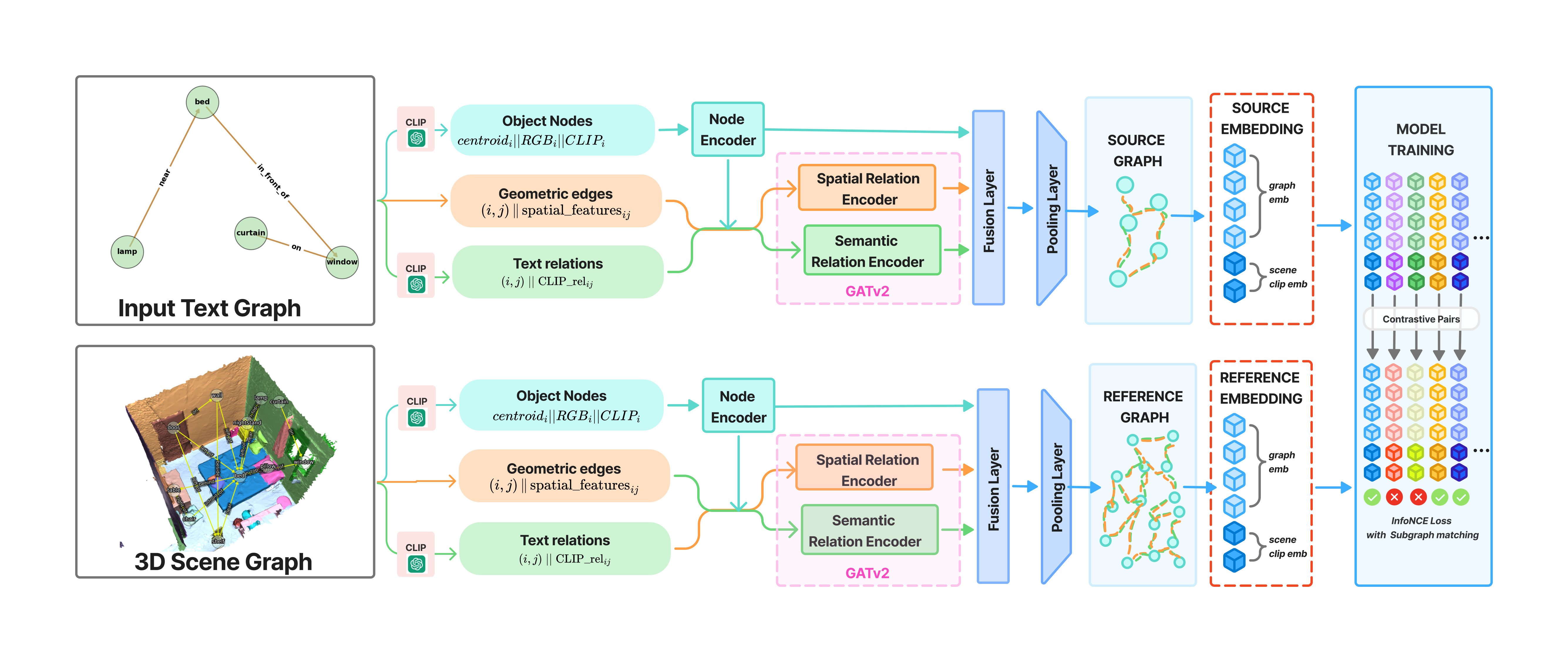}}
\vspace{-10mm}
\caption{\textbf{Coarse retrieval pipeline overview.} Given a natural language query, we first construct a scene graph representation where nodes encode object-level features (spatial attributes and CLIP embeddings) and edges capture relational constraints. This graph is processed through a dual-branch architecture that maintains separate pathways for geometric and semantic information before adaptive fusion. During training, an InfoNCE objective learns to align text-derived query graphs with 3D database graphs by learning discriminative embeddings. This simultaneously pulls together matching query-database pairs while pushing apart non-matching pairs. }
\label{fig:pipeline}
\end{figure*}

The first stage matches the text scene graph $G^t$ to the most compatible scene in $\mathcal{D}$ (\cref{fig:pipeline}). We use a dual-branch graph encoder with CLIP~\cite{radford2021CLIP} semantic features, extending~\cite{chen2024whereami,miao2024scenegraphloc}.

\myparagraph{Graph representation.}\label{sec:retrieval_repr}
Both 3D scene graphs and text scene graphs share the same node-edge structure, so we encode them identically.
Each node $v_i$ represents an object instance with semantic label $l_i$. We form a per-node feature by concatenating the object's 3D centroid $\mathbf{c}_i\in\mathbb{R}^3$, its mean RGB color $\mathbf{k}_i\in\mathbb{R}^3$, and a CLIP~\cite{radford2021CLIP} text embedding of the label:
\begin{equation}
\mathbf{f}_i = [\,\mathbf{c}_i \;\|\; \mathbf{k}_i \;\|\; \phi(l_i)\,] \in \mathbb{R}^{518},
\label{eq:node_feat}
\end{equation}
where $\phi(\cdot)\in\mathbb{R}^{512}$ is the CLIP text encoder (ViT-B/32). For text-derived graphs, where geometry and color are unknown, we zero-pad these fields and let the network rely on semantic and relational cues alone.

We also compute a scene-level CLIP descriptor by encoding a sentence that lists all unique object labels $\mathcal{L}(G)$ in the graph:
\begin{equation}
\mathbf{u}(G)= \phi(\text{``A room with } l_1, \ldots, l_K \text{''}),
\quad \{l_k\}_{k=1}^{K} = \mathcal{L}(G).
\label{eq:global_clip}
\end{equation}
This provides a coarse, holistic summary of what objects are present, complementing the fine-grained node-level features.

\myparagraph{Dual-branch encoder.}\label{sec:retrieval_encoder}
The encoder maps any graph to an embedding $\mathbf{z}(G)\in\mathbb{R}^{d}$ ($d{=}256$). 3D graphs carry both geometric layout and semantic relations; text graphs carry only relations. We handle this asymmetry with two branches, merged by a learned gate~\cite{miao2024scenegraphloc}.

Node features are first projected to the working dimension $d$ via an MLP:
\begin{equation}
\mathbf{x}_i = \mathrm{MLP}(\mathbf{f}_i) \in \mathbb{R}^{d}.
\label{eq:node_proj}
\end{equation}
The \emph{geometric branch} connects each node to its $k{=}5$ nearest spatial neighbors and propagates messages conditioned on their relative position and size:
\begin{equation}\begin{aligned}
\mathbf{e}^{\text{geom}}_{ij} &= [\,\Delta\mathbf{c}_{ij} \;\|\; \|\Delta\mathbf{c}_{ij}\|_2 \;\|\; r_i \;\|\; r_j \;\|\; \mathbf{0}_2\,],\\
\mathbf{g}_i& = \mathrm{GATv2}_{\text{geom}}(\{\mathbf{x}_j,\mathbf{e}^{\text{geom}}_{ij}\}_{j\in\mathcal{N}(i)}),
\label{eq:geom_branch}
\end{aligned}\end{equation}
where $\Delta\mathbf{c}_{ij}=\mathbf{c}_j-\mathbf{c}_i$ and $r_i,r_j$ are bounding radii. The \emph{relation branch} embeds each relation string with CLIP. Semantically similar relations (e.g., \textit{left\_of} and \textit{to the left of}) are thus naturally close in embedding space:
\begin{equation}
\mathbf{e}^{\text{rel}}_{ij} = \phi(r_{ij}), \quad
\mathbf{t}_i = \mathrm{GATv2}_{\text{rel}}(\{\mathbf{x}_j,\mathbf{e}^{\text{rel}}_{ij}\}_{j:(i,j)\in E}).
\label{eq:rel_branch}
\end{equation}
A learned gate $\boldsymbol{\alpha}_i = \sigma(\mathrm{MLP}([\mathbf{g}_i \| \mathbf{t}_i])) \in (0,1)^d$ fuses the two branches per node:
\begin{equation}
\mathbf{h}_i = \boldsymbol{\alpha}_i \odot \mathbf{g}_i + (1-\boldsymbol{\alpha}_i)\odot \mathbf{t}_i.
\label{eq:gated_fusion}
\end{equation}
For text graphs, where geometry is absent, the gate learns to rely primarily on the relation branch. Mean-pooling the node embeddings and concatenating the global descriptor yields the final graph embedding:
\begin{equation}
\mathbf{z}(G) = \mathrm{MLP}\!\biggl(\biggl[\frac{1}{|V|}\sum\nolimits_{i}\mathbf{h}_i \;\Big\|\; \mathbf{u}(G)\biggr]\biggr) \in \mathbb{R}^{d}.
\label{eq:final_embed}
\end{equation}

\myparagraph{Training.}\label{sec:retrieval_training}
We train with an InfoNCE contrastive loss~\cite{chen2024whereami} that pulls matching text-scene graph pairs together and pushes non-matching pairs apart. Each training pair consists of a sparse text graph $G^t$ and the corresponding dense 3D scene graph $G^s$; negatives are sampled from different scenes.

\myparagraph{Inference scoring.}\label{sec:retrieval_scoring}
At test time, we rank each database scene $G^s_i$ by a weighted combination of three complementary cues: the learned graph embedding similarity, the global CLIP descriptor similarity, and a simple label-overlap score:
\begin{equation}
\begin{aligned}
\mathrm{score}(G^t,G^s_i) =\;
&w_{\mathrm{emb}}\cos(\mathbf{z}(G^t),\mathbf{z}(G^s_i)) + w_{\mathrm{glob}}\cos(\mathbf{u}(G^t),\mathbf{u}(G^s_i)) \\
+\;&w_{\mathrm{jac}}\mathrm{F1}(\mathcal{L}(G^t),\mathcal{L}(G^s_i)),
\end{aligned}
\label{eq:retrieval_score}
\end{equation}
where $\mathrm{F1}$ measures the overlap of unique object labels and the weights $w_{\mathrm{emb}}$, $w_{\mathrm{glob}}$, $w_{\mathrm{jac}}$ are tuned on a validation split. Ranking by this score yields the retrieved list $\pi(T)$.

\subsection{Fine Localization}
\label{sec:fine_localization}

Once scene retrieval identifies the correct environment, the second stage estimates a precise pose within it. The key insight is that the observer must be standing at a position from which the mentioned objects are \emph{simultaneously visible and nearby}. We exploit this by matching text entities to 3D objects and then scoring a dense grid of floor positions via ray-casting against the scene mesh, counting how many matched objects are visible from each candidate.

\myparagraph{Object-level matching.}
We embed each node label and relationship predicate into $\mathbb{R}^{d}$ using Word2Vec~\cite{mikolov2013word2vec}.
Let $\mathbf{Q}\in\mathbb{R}^{|V^{t}|\times d}$ and $\mathbf{S}\in\mathbb{R}^{|V^{s}|\times d}$ collect the $\ell_2$-normalized node embeddings for query and scene nodes. We compute a cosine similarity matrix $\mathbf{M} = \hat{\mathbf{Q}}\,\hat{\mathbf{S}}^{\top}$. Each query node is greedily assigned to its best unused scene match above a similarity threshold.

\myparagraph{Candidate position grid.}
We sample candidate 2D floor positions $\mathbf{c}_k\in\mathbb{R}^2$ on a uniform grid over the walkable floor area at spacing $\Delta$. Each candidate represents a hypothesis for where the observer is standing. For visibility queries, we lift each candidate to 3D by adding a fixed eye height $h$ above the floor.

\myparagraph{Visibility scoring.}
For each candidate $\mathbf{c}_k$ and each matched object centroid $\boldsymbol{\mu}_o\in\mathbb{R}^3$, we cast a ray from the 3D eye position toward $\boldsymbol{\mu}_o$ on the scene mesh. The object is visible if the ray is not occluded. The score combines a visibility count with a proximity bonus that favors standing positions close to the mentioned objects:
\begin{equation}
s_k = \underbrace{\sum\nolimits_{o}\mathbb{I}[\mathrm{vis}(o,\mathbf{c}_k)]}_{\text{visibility count}} + w_d\underbrace{\sum\nolimits_{o:\,\mathrm{vis}(o,\mathbf{c}_k)} \exp\!\Big({-d_{ko}}/{\tau_d}\Big)}_{\text{proximity bonus}},
\end{equation}
where $d_{ko}$ is the horizontal distance from $\mathbf{c}_k$ to the object. Scores are converted to a probability distribution via softmax: $P(\mathbf{c}_k)\propto\exp(s_k/\tau)$.

\myparagraph{Orientation estimation.}
For the top-scoring positions, we estimate the viewing direction $\hat{\theta}_k\in\mathbb{S}^2$ by selecting the FoV frustum orientation that captures the most visible object directions, then taking the normalized mean of those directions.

\myparagraph{Final prediction.}
The output pose $\hat{\mathbf{p}}=(\hat{\mathbf{c}},\hat{\theta})$ is the weighted mean of posterior floor position paired with its estimated heading.

\begin{figure*}[t]
\vspace{-10mm}
\centering
\makebox[\textwidth][c]{\includegraphics[width=1.14\textwidth]{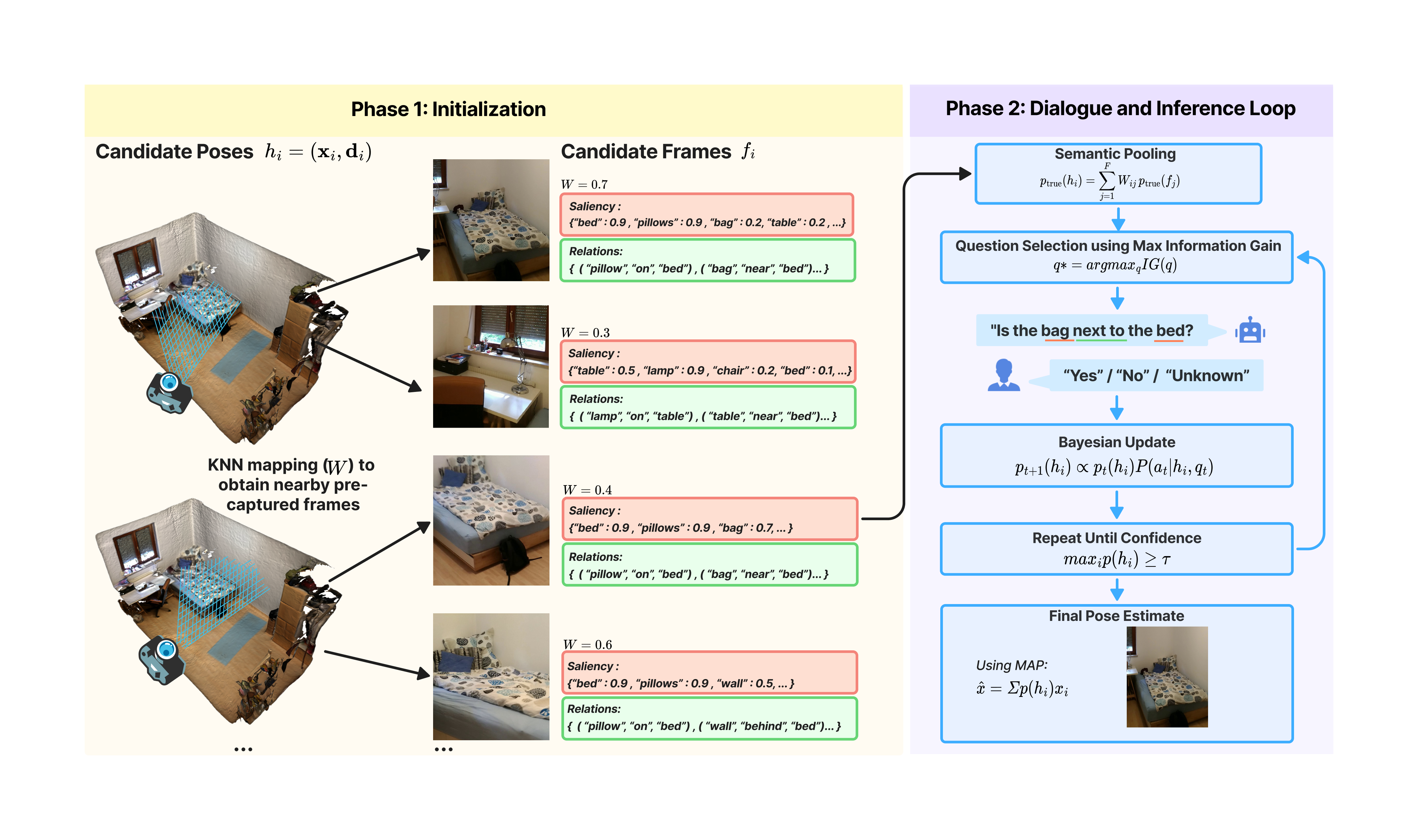}}
\vspace{-10mm}
\caption{\textbf{Dialog-based disambiguation overview.} \emph{Phase~1 (Initialization):} candidate floor poses $h_j{=}(\mathbf{x}_j,\mathbf{d}_j)$ from the fine-localization grid are mapped via KNN to nearby pre-captured reference frames $f_i$, each annotated with visible object labels, spatial relations, and a matching weight $w$. \emph{Phase~2 (Dialog and Inference Loop):} a semantic pooling step aggregates frame beliefs into a posterior $p(h_j)$. At each round, the system selects the yes/no question with maximum information gain (\eg ``Is the bag next to the bed?''), receives the user's answer, and updates the posterior via Bayes' rule. The loop repeats until the posterior concentrates above a confidence threshold $\tau$, at which point the final pose is returned as the mode of the posterior.}
\label{fig:dialog}
\end{figure*}

\begin{table}[t]
\centering
\small
\setlength{\tabcolsep}{5pt}
\caption{\textbf{Scene retrieval} Recall@$k$ (\%) on ScanScribe~\cite{chen2024whereami}. Each text query is matched against a pool of \textit{10 candidate scenes}. All methods encode text and 3D scene graphs into a shared embedding space and rank by similarity. Best in bold.}
\label{tab:whereami_tab1_scanscribe}
\resizebox{\columnwidth}{!}{%
\begin{tabular}{lcccc}
\toprule
Method & Top-1 & Top-2 & Top-3 & Top-5 \\
\midrule
Text2Pos (from scratch) & 11.63 $\pm$ 0.99 & 22.57 $\pm$ 1.16 & 33.03 $\pm$ 1.36 & 52.80 $\pm$ 1.67 \\
Text2Pos (fine-tuning)  & 10.30 $\pm$ 0.93 & 20.74 $\pm$ 1.32 & 31.65 $\pm$ 1.45 & 53.42 $\pm$ 1.68 \\
CLIP2CLIP              & 33.09 $\pm$ 0.21 & 52.53 $\pm$ 0.24 & 65.98 $\pm$ 0.21 & 82.69 $\pm$ 0.13 \\
Text2SGM (match-prob)  & 63.70 $\pm$ 1.85 & 81.40 $\pm$ 1.85 & 89.31 $\pm$ 1.42 & 95.71 $\pm$ 0.93 \\
Text2SGM (cos-sim)     & 68.27 $\pm$ 2.05 & 84.95 $\pm$ 1.62 & 91.65 $\pm$ 1.23 & 97.14 $\pm$ 0.76 \\
Text2SGM (ret-based)   & 68.61 $\pm$ 0.04 & 85.00 $\pm$ 0.02 & 91.57 $\pm$ 0.01 & 96.99 $\pm$ 0.00 \\
\textbf{LangLoc (ours)}  & \textbf{76.70 $\pm$ 4.58} & \textbf{90.40 $\pm$ 2.73} & \textbf{96.10 $\pm$ 1.58} & \textbf{98.90 $\pm$ 1.22} \\
\bottomrule
\end{tabular}}
\end{table}

\begin{table}[t]
\centering
\small
\setlength{\tabcolsep}{5pt}
\caption{\textbf{Scene retrieval} Recall@$k$ (\%) on ScanScribe~\cite{chen2024whereami}, retrieving from the full set of \textit{55 test scenes}. Unlike \cref{tab:whereami_tab1_scanscribe}, the candidate pool is not subsampled, making the task more challenging. We evaluate Top 5, 10, 20 and 30 scenes. {Since \cref{tab:whereami_tab1_scanscribe} and this table use different candidate-pool sizes (10 vs.\ 55) and different $k$ ranges, Recall@$k$ values are not directly comparable across the two tables.} Best in bold.}
\label{tab:whereami_tab2_scanscribe}
\resizebox{\columnwidth}{!}{%
\begin{tabular}{lcccc}
\toprule
Method & Top-5 & Top-10 & Top-20 & Top-30 \\
\midrule
Text2Pos (from scratch) & 10.27 $\pm$ 0.83 & 21.24 $\pm$ 1.17 & 39.36 $\pm$ 1.43 & 57.30 $\pm$ 1.48 \\
Text2Pos (fine-tuning)  &  9.21 $\pm$ 0.92 & 18.27 $\pm$ 1.24 & 39.27 $\pm$ 1.61 & 58.63 $\pm$ 1.60 \\
CLIP2CLIP              & 33.26 $\pm$ 0.21 & 53.47 $\pm$ 0.23 & 73.76 $\pm$ 0.18 & 86.23 $\pm$ 0.11 \\
Text2SGM (match-prob)  & 70.24 $\pm$ 1.93 & 85.71 $\pm$ 1.38 & 93.36 $\pm$ 1.06 & 97.25 $\pm$ 0.65 \\
Text2SGM (cos-sim)     & 76.34 $\pm$ 2.06 & 87.83 $\pm$ 1.55 & 95.40 $\pm$ 0.89 & 98.24 $\pm$ 0.57 \\
Text2SGM (ret-based)   & 76.29 $\pm$ 0.16 & 87.77 $\pm$ 0.10 & 95.34 $\pm$ 0.05 & 98.18 $\pm$ 0.02 \\
\textbf{LangLoc (ours)} & \textbf{83.30 $\pm$ 3.74} & \textbf{91.60 $\pm$ 3.01} & \textbf{97.10 $\pm$ 1.51} & \textbf{98.80 $\pm$ 0.87} \\
\bottomrule
\end{tabular}}
\end{table}

\begin{table}[t]
\centering
\small
\setlength{\tabcolsep}{5pt}
\caption{\textbf{Scene retrieval} Recall@$k$ (\%) on ScanScribe~\cite{chen2024whereami} using \textit{LLM-generated queries} from scene images instead of scene-graph-derived text. This reduces lexical overlap between query terms and database object labels, testing generalization to natural phrasing. Best in bold.}
\label{tab:whereami_tab4_images}
\resizebox{\columnwidth}{!}{%
\begin{tabular}{lcccc}
\toprule
Method & Top-1 & Top-2 & Top-3 & Top-5 \\
\midrule
CLIP2CLIP          & 33.07 $\pm$ 0.21 & 52.52 $\pm$ 0.23 & 66.00 $\pm$ 0.21 & 82.78 $\pm$ 0.13 \\
Text2SGM (cos-sim) & 34.22 $\pm$ 1.77 & 56.67 $\pm$ 1.70 & 68.78 $\pm$ 2.35 & 82.11 $\pm$ 1.23 \\
\midrule
LangLoc (ours)     & \textbf{59.5 $\pm$ 5.26} &  \textbf{76.4 $\pm$ 4.94} & \textbf{87.8 $\pm$3.12} & \textbf{96.2 $\pm$ 2.27} \\
\bottomrule
\end{tabular}}
\end{table}

\subsection{Dialog-based Disambiguation}
\label{sec:dialog}

When several spatially distinct viewpoints receive similar scores, a single description may not uniquely determine the pose. A room may, for instance, contain two similar seating areas, each consistent with ``a sofa facing a TV''. To resolve such ambiguities, we introduce a dialog module (\cref{fig:dialog}) that asks targeted yes/no questions and uses the answers to iteratively narrow the set of plausible viewpoints via Bayesian posterior updates.

\myparagraph{Frame-level posterior.}
We discretize the pose space into several reference frames $\{(\mathbf{t}_j,\theta_j)\}_{j=1}^{F}$, where $\mathbf{t}_j\in\mathbb{R}^2$ is the floor position and $\theta_j$ is the viewing direction. Each frame is annotated with visible labels $\mathcal{L}_j$ and spatial relations $\mathcal{R}_j$. In practice, we first \emph{pool} a compact set of representative frames (e.g., the most frequently matched frames under a candidate-to-frame KNN mapping), and perform dialog only over this pooled set.
The fine-localization belief over candidate poses is mapped to frames via a distance-weighted soft assignment matrix $W$, yielding the initial frame belief:
\begin{equation}
p_0(j) \propto (W^\top p_C)_j,
\label{eq:init_frame_posterior}
\end{equation}
where $p_C$ denotes the (normalized) prior over candidate poses.

\myparagraph{Bayesian update.}
Each round asks a question $q$ and receives $a\in\{\text{yes},\text{no},\text{unknown}\}$. We use \emph{label} questions (object visibility) and \emph{relation} questions (spatial configuration). For each frame $j$, we compute a soft truth probability $p_{\text{true}}(j)\in[0,1]$ and answerability $p_{\text{ans}}(j)\in[0,1]$: for labels, $p_{\text{true}}(j)$ comes from visibility/salience in frame $j$; for relations, $p_{\text{true}}(j)$ is determined by membership in $\mathcal{R}_j$ and $p_{\text{ans}}(j)$ by the involved objects. A reliability $\alpha$ flips yes/no with probability $(1-\alpha)$, and we allocate explicit mass to \texttt{unknown} that increases as $p_{\text{ans}}(j)$ decreases. We update the posterior by
\begin{equation}
p(j) \leftarrow \frac{p(j)\,\mathcal{L}(a\mid j)}{\sum_{j'}p(j')\,\mathcal{L}(a\mid j')}.
\label{eq:bayes_update}
\end{equation}

\myparagraph{Question selection.}
We greedily select an informative question under the current posterior. By default, we maximize expected information gain (including \texttt{unknown}):
\begin{equation}
s(q)= H(p)-\sum\nolimits_{a\in\{\text{y},\text{n},\text{u}\}} p(a\mid q)\,H\!\bigl(p(\cdot\mid a,q)\bigr), \qquad
p(a\mid q)=\sum_j p(j)\,\mathcal{L}(a\mid j,q).
\label{eq:ig_score}
\end{equation}
As a lightweight alternative, we support the balanced split heuristic
\begin{equation}
s_{\text{bal}}(q) = -\bigl|p_{\text{yes}}(q) - \tfrac{1}{2}\bigr|, \qquad
p_{\text{yes}}(q)=\sum\nolimits_j p(j)\,y_j,
\label{eq:question_score}
\end{equation}
with $y_j$ indicating whether frame $j$ implies a ``yes''. Relation questions are preferred when available; we filter out questions that are too unbalanced or insufficiently answerable under $p(\cdot)$. For label questions, we apply an IDF-based downweighting to avoid frequently occurring, low-discriminative labels.

\myparagraph{Pose estimation.}
After $R$ dialog rounds, the final pose is the mode of the posterior:
\begin{equation}
\hat{\mathbf{c}} = \mathbf{t}_{j^*} \in \mathbb{R}^2,
\qquad
\hat{\theta} = \theta_{j^*} \in \mathbb{S}^2,
\qquad
j^* = \arg\max_{j}\; p(j).
\label{eq:dialog_pose}
\end{equation}
We report \emph{Pos.}~(m) as the error of $\hat{\mathbf{c}}$ in
(Tab.~\ref{tab:fine_loc_two_datasets}).


\section{LangLoc Dataset}
\label{sec:dataset}

Existing 3D vision-language datasets provide object-centric grounding~\cite{chen2020scanrefer,achlioptas2020referit_3d} or scene-level captions~\cite{zhu20233dvista}, but not pose-indexed egocentric descriptions suitable for localization. We build such a benchmark by extending 3RScan~\cite{Wald2019RIO} with the pose-indexed text descriptions over $1{,}300{+}$ indoor scans, with $13{,}000{+}$ pose-indexed natural-language descriptions.

\myparagraph{Keyframe selection.}
Raw RGB-D sequences contain many blurry, redundant, or uninformative frames. We select a compact set of high-quality keyframes per scene in three steps. First, each frame is scored by an image quality model~\cite{agnolucci2025qualiCLIP} and low-quality frames are discarded. Second, we render the scene mesh to determine which objects are visible from each surviving frame and compute pairwise spatial relations (e.g., \textit{left\_of}, \textit{above}) in camera coordinates. Third, we apply a two-stage determinantal point process (DPP)~\cite{kulesza2012determinantal}: Stage~1 selects semantically informative frames using a quality score that rewards object diversity and geometric complexity; Stage~2 enforces spatial diversity by penalizing frames with overlapping positions, viewing directions, and visibility masks.

\begin{table*}[t]

\caption{Fine localization on 100-scene subsets of two datasets. Given the correct scene and a text description, estimate the observer's 2D floor position and heading. \emph{Midpoint}: floor centroid. \emph{VLM}: Qwen2.5-VL-2B~\cite{qwen2vl} with top-down rendering and query (zero-shot). \emph{Top-10 Pos.}: minimum position error among the 10 highest-scoring grid cells. \emph{3D IoU}: frustum intersection-over-union between predicted and ground-truth views ($\uparrow$). For the dialog setting, answers are obtained in a controlled protocol: by a human annotator on 3RScan and by Qwen2.5-VL-2B using ground-truth reference-frame metadata on ScanNet. Thus, the dialog rows should be interpreted as controlled dialog evaluations rather than full real-user studies. All other metrics: lower is better. Best in bold, second best underlined.}

\vspace{-4mm}
\label{tab:fine_loc_two_datasets}
\centering

\begin{subtable}[t]{0.495\textwidth}
\centering
\caption{\textbf{3RScan~\cite{Wald2019RIO}}}
\footnotesize
\setlength{\tabcolsep}{3pt}
\renewcommand{\arraystretch}{1.12}
\resizebox{\linewidth}{!}{%
\begin{tabular}{l
                S[table-format=1.3] S[table-format=1.3]
                S[table-format=1.3] S[table-format=1.3]
                S[table-format=2.2] S[table-format=2.2]
                S[table-format=1.3]}
\toprule
& \multicolumn{2}{c}{\textbf{Top-10 Pos.\,(m)} $\downarrow$}
& \multicolumn{2}{c}{\textbf{Pos.\,(m)} $\downarrow$}
& \multicolumn{2}{c}{\textbf{Angle\,(deg)} $\downarrow$}
& {\textbf{3D IoU} $\uparrow$} \\
\cmidrule(lr){2-3}\cmidrule(lr){4-5}\cmidrule(lr){6-7}\cmidrule(lr){8-8}
\textbf{Method} & {\textbf{Mean}} & {\textbf{Med.}}
                & {\textbf{Mean}} & {\textbf{Med.}}
                & {\textbf{Mean}} & {\textbf{Med.}}
                & {\textbf{Mean}} \\
\midrule
Midpoint
& {--} & {--}
& \underline{1.416} & \underline{1.347}
& {--} & {--}
& {--} \\
VLM
& {--} & {--}
& 1.582 & 1.420
& 85.48 & 85.54
& 0.062 \\
LangLoc w/o dialog
& \underline{1.037} & \underline{0.941}
& 1.712 & 1.551
& \underline{46.07} & \underline{37.24}
& \underline{0.172} \\
LangLoc w/\phantom{o} dialog
& \textbf{0.983} & \textbf{0.890}
& \textbf{0.926} & \textbf{0.799}
& \textbf{39.52} & \textbf{33.37}
& \textbf{0.342} \\
\bottomrule
\end{tabular}}
\end{subtable}
\hfill
\begin{subtable}[t]{0.495\textwidth}
\centering
\caption{\textbf{ScanNet~\cite{dai2017scannet}}}
\footnotesize
\setlength{\tabcolsep}{3pt}
\renewcommand{\arraystretch}{1.12}
\resizebox{\linewidth}{!}{%
\begin{tabular}{l
                S[table-format=1.3] S[table-format=1.3]
                S[table-format=1.3] S[table-format=1.3]
                S[table-format=2.2] S[table-format=2.2]
                S[table-format=1.3]}
\toprule
& \multicolumn{2}{c}{\textbf{Top-10 Pos.\,(m)} $\downarrow$}
& \multicolumn{2}{c}{\textbf{Pos.\,(m)} $\downarrow$}
& \multicolumn{2}{c}{\textbf{Angle\,(deg)} $\downarrow$}
& {\textbf{3D IoU} $\uparrow$} \\
\cmidrule(lr){2-3}\cmidrule(lr){4-5}\cmidrule(lr){6-7}\cmidrule(lr){8-8}
\textbf{Method} & {\textbf{Mean}} & {\textbf{Med.}}
                & {\textbf{Mean}} & {\textbf{Med.}}
                & {\textbf{Mean}} & {\textbf{Med.}}
                & {\textbf{Mean}} \\
\midrule
Midpoint
& {--} & {--}
& \underline{1.279} & \underline{1.098}
& {--} & {--}
& {--} \\
VLM
& {--} & {--}
& 1.382 & 1.099
& 93.96 & 91.27
& 0.030 \\
LangLoc w/o dialog
& \underline{1.254} & \underline{1.065}
& 1.676 & 1.314
& \underline{42.67} & \underline{34.66}
& \underline{0.236} \\
LangLoc w/\phantom{o} dialog
& \textbf{0.532} & \textbf{0.210}
& \textbf{0.593} & \textbf{0.069}
& \textbf{23.77} & \phantom{1}\textbf{5.12}
& \textbf{0.593} \\
\bottomrule
\end{tabular}}
\end{subtable}

\end{table*}

\myparagraph{Description generation.}
For each selected keyframe, the visible object list and spatial relations are assembled into a structured prompt for an LLM, which produces a natural-language scene description grounded at that viewpoint. These automatic descriptions are supplemented by human annotations collected via a dedicated annotation interface to form the final training and evaluation labels. Full pipeline details and hyperparameters are provided in the supp.\ material.

\section{Experiments}
\label{sec:experiments}

We evaluate scene retrieval (\cref{sec:exp_retrieval}) and fine localization with and without dialog (\cref{sec:exp_fine_loc}).
For retrieval, we use the ScanScribe benchmark~\cite{chen2024whereami} built on 3RScan~\cite{Wald2019RIO} with 3DSSG~\cite{3DSSG2020} annotations and follow its standard train/val/test split.
For fine localization, we evaluate on the LangLoc dataset (\cref{sec:dataset}), built on 3RScan, and on ScanNet~\cite{dai2017scannet}.

\myparagraph{Implementation details.}
Text descriptions are parsed into scene graphs with GPT-4o-mini.
Node features concatenate centroid coordinates (3D), color attributes (3D), and CLIP ViT-B/32 embeddings (512D), totaling 518 dimensions.
Scene-level CLIP embeddings are computed from object label sets and fused with graph embeddings via a two-layer MLP with LayerNorm.
The dual-branch GATv2 encoder is trained for {70} epochs using AdamW (learning rate {$5\times10^{-4}$}, weight decay $10^{-4}$) with a contrastive loss at temperature {0.07}, 50\% negative sampling, and gradient clipping (max norm 1.0).
For fine localization, we sample a dense floor grid at spacing $\Delta{=}0.25$\,m and eye height $h{=}1.6$\,m, and score each cell by single-ray visibility casting against the scene mesh.

\subsection{Scene Retrieval}
\label{sec:exp_retrieval}

We report Recall@$k$ under the three evaluation protocols of~\cite{chen2024whereami}. Baselines are taken from~\cite{chen2024whereami}: Text2Pos~\cite{kolmet2022text2pos} learns a joint text--point-cloud embedding, CLIP2CLIP matches CLIP embeddings of text and rendered views, and Text2SGM performs scene-graph matching (match-prob, cos-sim, ret-based).

\myparagraph{10-scene pool (\cref{tab:whereami_tab1_scanscribe}).}
Each query is matched against 10 candidate scenes. LangLoc achieves 76.7\% Top-1 recall, exceeding the strongest Text2SGM variant (68.6\%) by $+8.1$\,pp, and reaches 98.9\% at Top-5. The gains across $k$ suggest that our dual-branch CLIP-based encoder with gated fusion learns more discriminative scene embeddings than single-branch alternatives.

\myparagraph{Full test set (\cref{tab:whereami_tab2_scanscribe}).}
Retrieval over all 55 test scenes is harder ($5.5\times$ larger pool). LangLoc reaches 83.3\% Top-5 and 91.6\% Top-10, outperforming Text2SGM by 7 and 4 points, respectively. Text2Pos drops below 10\% at Top-5, while CLIP2CLIP improves to 33.3\% Top-5 but remains far below graph-based methods, highlighting the benefit of object-level structure and spatial relations.

\myparagraph{LLM-generated queries (\cref{tab:whereami_tab4_images}).}
To test robustness beyond graph-derived text, we generate queries with GPT-4o-mini from scene images and detected objects, reducing lexical overlap and increasing phrasing variability. {LangLoc obtains 59.5\% Top-1, a $+25$\,pp gain over Text2SGM (34.2\%) and $+26$\,pp over CLIP2CLIP (33.1\%); at Top-5 it reaches 96.2\%, indicating the correct scene is almost always in the top candidates under domain shift.}

\subsection{Fine Localization}
\label{sec:exp_fine_loc}

Given the retrieved scene, we evaluate how accurately LangLoc estimates the observer's 2D floor position $\hat{\mathbf{c}}$ and heading $\hat{\theta}$.
We compare against two baselines that require no task-specific training.

\myparagraph{Baselines.}
The \emph{midpoint} baseline places the observer at the centroid of the floor bounding box; it uses no language input and does not predict a heading, so angular error and IoU are not applicable.
The \emph{VLM} baseline provides Qwen2.5-VL-2B~\cite{qwen2vl} with a top-down rendering of the scene and the text query, and asks it to predict pixel coordinates and a heading in a zero-shot setting.
The predicted pixel locations are unprojected to 3D via the known intrinsics and floor-plane intersection.

\myparagraph{Metrics.}
We report four metrics.
\emph{Position error}: Euclidean distance $\|\hat{\mathbf{c}} - \mathbf{c}^\star\|_2$ in meters over 2D locations.
\emph{Top-$k$ position error}: minimum distance among the $k$ highest-scoring grid cells, providing a softer measure for multi-modal posteriors.
\emph{Angular error}: geodesic angle $\arccos\!\left(\operatorname{clip}(\hat{\theta}^{\top}\theta^\star,-1,1)\right)$ between the predicted and ground-truth unit direction vectors, reported in degrees.
\emph{3D IoU}: frustum intersection-over-union between the predicted and ground-truth viewing frustums on the floor plane (higher is better).

\myparagraph{Evaluation data.}
Because the dialog-based variant requires human annotations -- each query involves multiple rounds of interactive disambiguation -- we evaluate on 100 randomly selected scenes from each dataset.
Annotating dialog turns is time-consuming, so we reserve the full-scale evaluation (all 1,300 scans of the LangLoc dataset) for the non-dialog variant and report it separately in \cref{tab:fine_loc}; complete per-dataset results are provided in the supplementary material.

\myparagraph{Results on 100-scene subsets (\cref{tab:fine_loc_two_datasets}).}
On the 3RScan split, LangLoc without dialog already halves the angular error compared to the VLM baseline (46.1$^\circ$ mean vs.\ 85.5$^\circ$), demonstrating that ray-cast visibility scoring yields a meaningful heading estimate.
The midpoint does not predict a heading (marked ``--'' in the table).
Adding dialog further reduces the median position error from 1.55\,m to 0.80\,m -- a 49\% improvement -- and the mean angular error from 46.1$^\circ$ to 39.5$^\circ$.
The IoU nearly doubles from 0.172 to 0.342, as the dialog module narrows the posterior and aligns the predicted frustum with the ground truth.
The VLM achieves only 0.062 IoU despite predicting a heading, since its orientations are essentially random (${\sim}85^\circ$ error).

\begin{figure*}[t]
    \centering


    \includegraphics[width=0.25\textwidth]{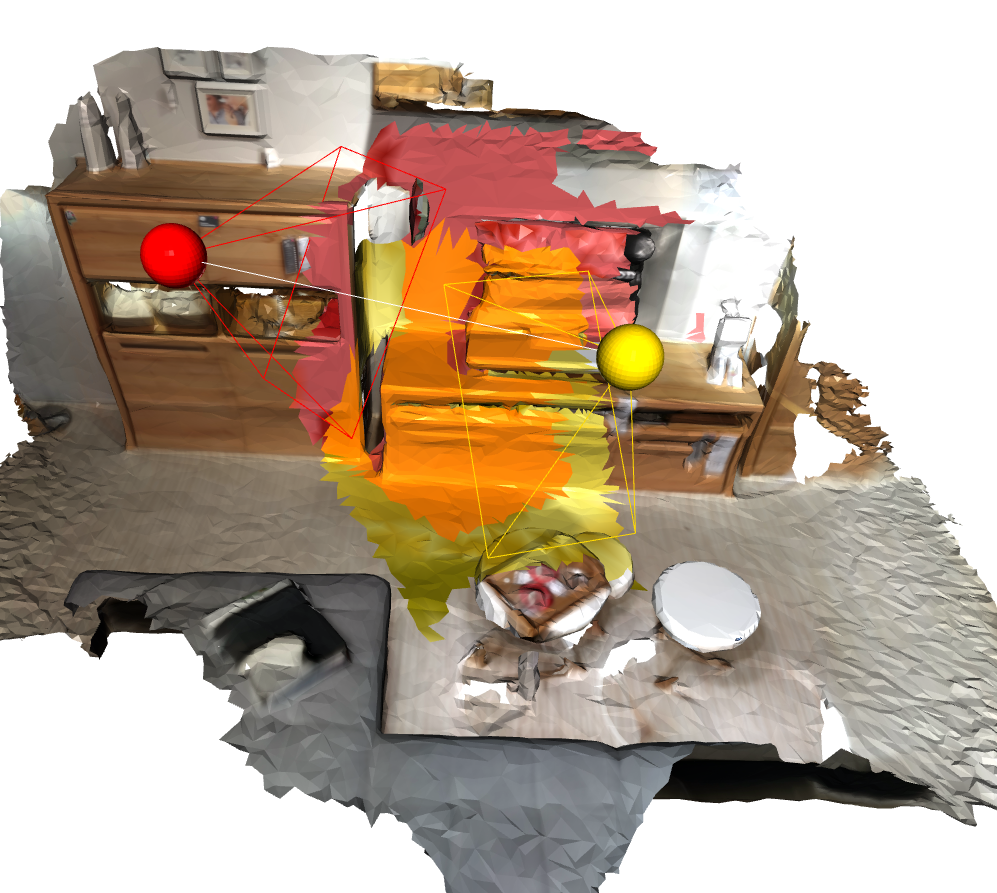}\hfill
    \includegraphics[width=0.25\textwidth]{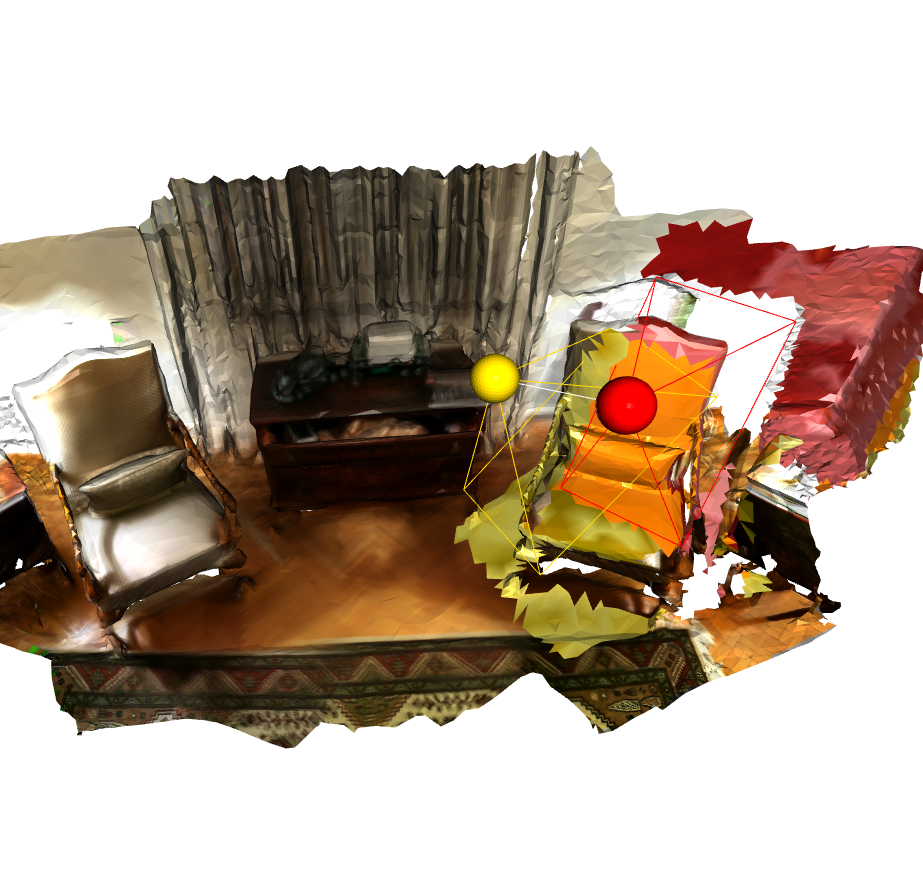}\hfill
    \includegraphics[width=0.25\textwidth]{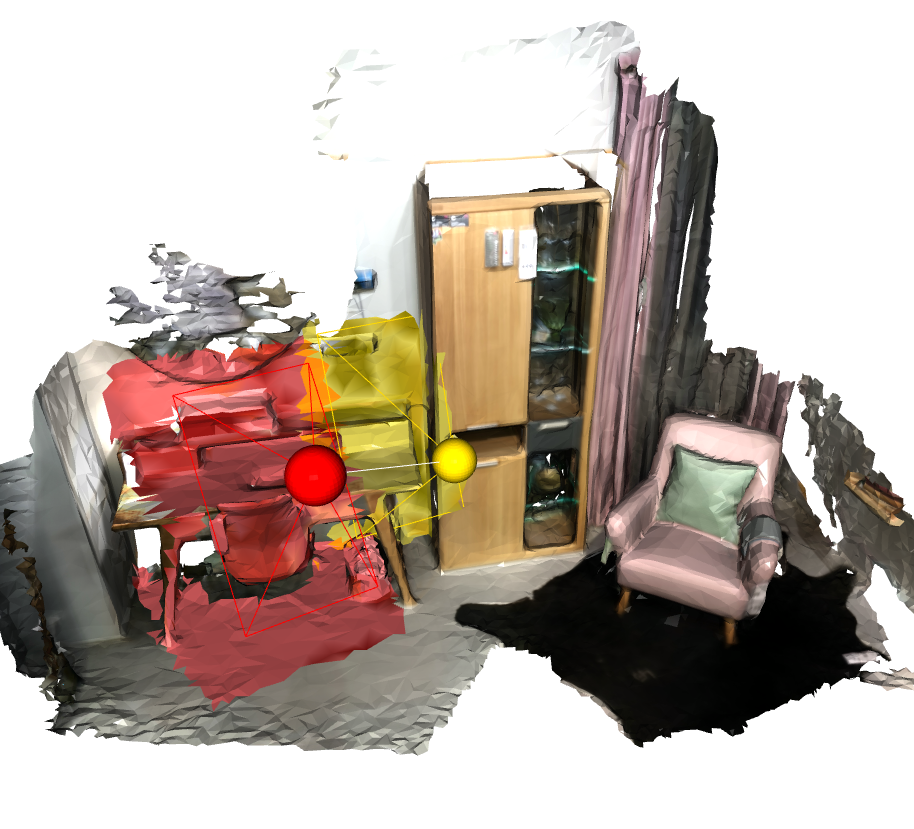}\hfill
    \includegraphics[width=0.25\textwidth]{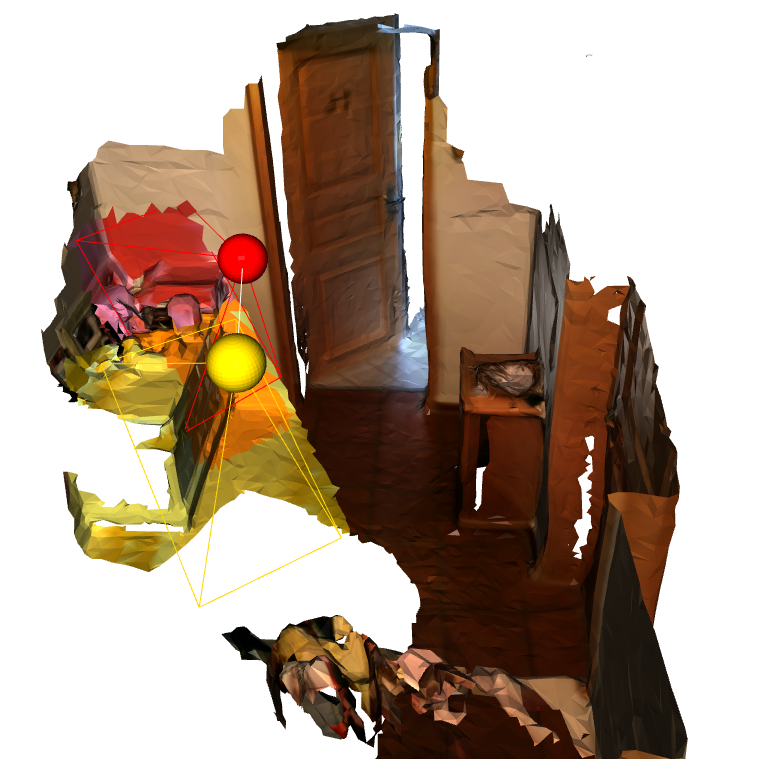}\hfill

\caption{\textbf{Qualitative frustum overlap on two datasets.} We visualize ground-truth frustum (red), predicted frustum (yellow), and their overlap (orange).}
    \label{fig:scannet_3rscan_qual}
\end{figure*}

An instructive pattern emerges in the raw position error.
The midpoint baseline achieves a seemingly competitive 1.42\,m mean despite using \emph{no} language at all.
This is an artifact of small indoor rooms: the floor centroid is mechanically close to every point in a compact space, but it carries no practical value -- it conveys neither a meaningful within-room position nor a viewing direction.
Therefore, the midpoint baseline should not be interpreted as a meaningful localization result, since it provides neither orientation nor viewpoint-specific evidence, unlike LangLoc's angular, frustum-overlap, and posterior-based estimates.

LangLoc w/o dialog reports a higher raw position error (1.71\,m) because it commits to a specific floor region near the described objects, which can be wrong when several object clusters match the description.
This suggests that the main residual ambiguity is not heading estimation, where LangLoc is already far stronger than the VLM, but choosing among multiple spatially separated pose clusters that satisfy the same description.
The dialog module targets exactly this failure mode by asking discriminative questions across clusters, thereby collapsing the posterior toward the correct viewpoint.

The Top-10 metric disambiguates the two behaviors: the correct position is almost always among the high-scoring cells, reaching 1.04\,m (Top-10 mean) -- well below the midpoint's raw error.
With dialog, LangLoc achieves 0.93\,m mean and 0.80\,m median position error, outperforming the midpoint on every metric.

On ScanNet, which features different room layouts and annotation conventions, LangLoc w/o dialog generalizes well: it achieves a Top-10 mean of 1.25\,m, an angular error of 42.7$^\circ$, and an IoU of 0.236 -- comparable to the 3RScan split.
The VLM again produces near-random headings (${\sim}94^\circ$) and near-zero IoU (0.030).
Adding the dialog on ScanNet yields even larger gains than on 3RScan: the median position error drops from 1.31\,m to 0.07\,m, the median angular error from 34.7$^\circ$ to 5.1$^\circ$, and the IoU rises from 0.236 to 0.593.

Two factors compound this: ScanNet frames contain more visible objects on average (6 vs.\ 4), making each dialog question more discriminative, and its rooms are less spatially ambiguous, causing the posterior to collapse to a single mode in over 40\% of scenes and snap the MAP estimate below grid spacing.

Here, Pos. and Top-10 Pos. reflect different estimators: the former
is the mode of the posterior ($j^* = \arg\max_j p(j)$, always on a
grid cell), whereas the latter is the minimum distance among the ten
highest-scoring grid cells, providing a softer measure when the
posterior is multi-modal. The Top-10 mean reaches 0.53\,m,
confirming that the Bayesian posterior concentrates sharply around
the correct pose after a few clarification rounds.

\myparagraph{Human dialog pilot.}
To test whether the controlled dialog gains transfer to real interaction, we additionally run a 10-scene pilot with human-written descriptions and human-typed dialog answers.
LangLoc with human dialog improves over single-shot LangLoc and approaches a human asked to localize from the same description (these human baselines are in the last row of \cref{tab:human_dialog_pilot}).
While the human annotator better estimates the heading directions, LangLoc with the dialog achieves the lowest position errors. 

\begin{table}[t]
\caption{\textbf{Human dialog pilot} on 10 scenes. Unlike the controlled dialog rows in \cref{tab:fine_loc_two_datasets}, this pilot uses human-written descriptions and human-typed dialog answers. The \emph{Human} row reports a person localizing from the same description. 3D IoU measures frustum intersection-over-union between predicted and ground-truth views; higher is better. All other metrics are lower-is-better.}
\label{tab:human_dialog_pilot}
\centering
\scriptsize
\setlength{\tabcolsep}{2pt}
\renewcommand{\arraystretch}{1.0}
\resizebox{\columnwidth}{!}{%
\begin{tabular}{@{}lccccccc@{}}
\toprule
& \multicolumn{2}{c}{\textbf{Top-10 Pos. (m)} $\downarrow$}
& \multicolumn{2}{c}{\textbf{Pos. (m)} $\downarrow$}
& \multicolumn{2}{c}{\textbf{Angle (deg)} $\downarrow$}
& \textbf{3D IoU} $\uparrow$ \\
\cmidrule(lr){2-3}\cmidrule(lr){4-5}\cmidrule(lr){6-7}\cmidrule(lr){8-8}
\textbf{Method} & Mean & Med. & Mean & Med. & Mean & Med. & Mean \\
\midrule
Midpoint & -- & -- & 0.91 & 0.83 & -- & -- & -- \\
Qwen & -- & -- & 1.26 & 0.85 & 78.68 & 75.38 & 0.03 \\
LangLoc & 1.16 & 0.96 & 1.74 & 1.19 & 49.93 & 54.08 & 0.21 \\
LangLoc + Human dialog & \textbf{0.61} & \textbf{0.24} & \textbf{0.66} & \textbf{0.36} & \underline{45.68} & \underline{40.34} & \underline{0.25} \\
\midrule
Human & -- & -- & \underline{0.70} & \underline{0.51} & \textbf{20.37} & \textbf{15.31} & \textbf{0.39} \\
\bottomrule
\end{tabular}}
\end{table}

\begin{table}[t]
\caption{\textbf{Fine localization} on the 1,300 scenes of the full LangLoc dataset. Given the correct scene and a text description, estimate the observer's 2D floor position and heading. \emph{Midpoint}: floor centroid (no language). \emph{VLM}: Qwen2.5-VL-2B~\cite{qwen2vl} with top-down rendering and query (zero-shot). \emph{Top-10 Pos.}: minimum position error among the 10 highest-scoring candidate locations. \emph{3D IoU}: frustum intersection-over-union between predicted and ground-truth views ($\uparrow$). All other metrics: lower is better. Best in bold, second best underlined.}
\label{tab:fine_loc}
\centering
\footnotesize
\setlength{\tabcolsep}{3.5pt}
\renewcommand{\arraystretch}{1.15}
\resizebox{\columnwidth}{!}{%
\begin{tabular}{l
                S[table-format=1.3] S[table-format=1.3]
                S[table-format=1.3] S[table-format=1.3]
                S[table-format=2.2] S[table-format=2.2]
                S[table-format=1.3]}
\toprule
& \multicolumn{2}{c}{\textbf{Top-10 Pos.\,(m)} $\downarrow$}
& \multicolumn{2}{c}{\textbf{Pos.\,(m)} $\downarrow$}
& \multicolumn{2}{c}{\textbf{Angle\,(deg)} $\downarrow$}
& {\textbf{3D IoU} $\uparrow$} \\
\cmidrule(lr){2-3}\cmidrule(lr){4-5}\cmidrule(lr){6-7}\cmidrule(lr){8-8}
\textbf{Method} & {\textbf{Mean}} & {\textbf{Med.}}
                & {\textbf{Mean}} & {\textbf{Med.}}
                & {\textbf{Mean}} & {\textbf{Med.}}
                & {\textbf{Mean}} \\
\midrule
Midpoint
& {--} & {--}
& \textbf{1.369} & \textbf{1.259}
& {--} & {--}
& {--} \\
VLM (Qwen2.5-VL)
& {--} & {--}
& 1.538 & 1.350
& \underline{84.70} & \underline{81.70}
& \underline{0.018} \\
LangLoc w/o dialog
& \textbf{1.153} & \textbf{0.951}
& \underline{1.534} & \underline{1.308}
& \textbf{46.85} & \textbf{39.80}
& \textbf{0.147} \\
\bottomrule
\end{tabular}}
\end{table}

\myparagraph{Full-scale evaluation (\cref{tab:fine_loc}).}
On the complete LangLoc dataset (1,300 scans, 13k+ descriptions), LangLoc w/o dialog achieves a Top-10 median position error of 0.95\,m and a median angular error of 39.8$^\circ$, consistent with the 100-scene subset and confirming that LangLoc scales to the full benchmark.
Midpoint again achieves a low raw position error (1.26\,m median vs.\ 1.31\,m for LangLoc) due to the small room sizes, but this remains a vacuous baseline: it provides neither a meaningful within-room position nor any viewing direction.
LangLoc's 3D IoU of 0.147 is an order of magnitude higher than the VLM's 0.018, confirming that the method recovers both accurate position and orientation from language alone.

\section{Conclusion}
\label{sec:conclusion}

We presented LangLoc, the first pipeline for fine-grained indoor localization that estimates a 2D position and heading from natural language text description -- without any image -- by combining a dual-branch GATv2 retrieval encoder, visibility-based floor-grid pose scoring, and a Bayesian dialog module for disambiguation.
On the proposed LangLoc dataset (13k+ descriptions over 1,300+ scans) and on ScanNet, it achieves approximately 1\,m median position accuracy and meaningful headings, substantially outperforming geometry-only and zero-shot VLM baselines.
Interactive dialog yields further large gains -- reducing the ScanNet median error to 7\,cm and 5$^\circ$ -- showing that a few targeted yes/no questions resolve much of the spatial ambiguity inherent in a single description.

\myparagraph{Limitations and future work.}
The current pipeline operates on pre-built 3D scene graphs with known object labels and relationships. Extending it to open-vocabulary or incrementally constructed graphs would broaden applicability to environments without prior annotation.
Performance degrades in large, cluttered scenes where many viewpoints share similar object configurations -- a challenge that richer spatial reasoning or multi-turn dialog strategies could address.
Finally, while language queries are inherently privacy-preserving, the method still requires access to a detailed 3D model of the environment; relaxing this assumption, \eg by localizing against coarser floor plans or schematic maps, is a promising direction for practical deployment.


%
%
    \bibliographystyle{splncs04}
\bibliography{main}

@String(CVPR  = {IEEE Conf. Comput. Vis. Pattern Recog.})

@String(ICCV  = {Int. Conf. Comput. Vis.})

@String(ECCV  = {Eur. Conf. Comput. Vis.})

@String(NeurIPS = {Adv. Neural Inform. Process. Syst.})

@String(ICLR  = {Int. Conf. Learn. Represent.})

@String(CVPRW = {IEEE Conf. Comput. Vis. Pattern Recog. Worksh.})

@String(AAAI  = {AAAI})

@String(CVPR  = {CVPR})

@String(ICCV  = {ICCV})

@String(ECCV  = {ECCV})

@String(NeurIPS = {NeurIPS})

@String(ICLR  = {ICLR})

@String(CVPRW = {CVPRW})

@inproceedings{miao2024scenegraphloc,
  title={Scenegraphloc: Cross-modal coarse visual localization on 3d scene graphs},
  author={Miao, Yang and Engelmann, Francis and Vysotska, Olga and Tombari, Federico and Pollefeys, Marc and Bar{\'a}th, D{\'a}niel B{\'e}la},
  booktitle={European Conference on Computer Vision},
  pages={127--150},
  year={2024},
  organization={Springer}
}

@inproceedings{sarkar2023sgaligner,
  title={Sgaligner: 3d scene alignment with scene graphs},
  author={Sarkar, Sayan Deb and Miksik, Ondrej and Pollefeys, Marc and Barath, Daniel and Armeni, Iro},
  booktitle={Proceedings of the IEEE/CVF International Conference on Computer Vision},
  pages={21927--21937},
  year={2023}
}

@inproceedings{chen2024whereami,
  title={“where am i?” scene retrieval with language},
  author={Chen, Jiaqi and Barath, Daniel and Armeni, Iro and Pollefeys, Marc and Blum, Hermann},
  booktitle={European Conference on Computer Vision},
  pages={201--220},
  year={2024},
  organization={Springer}
}

@inproceedings{armeni20193d,
  title={3d scene graph: A structure for unified semantics, 3d space, and camera},
  author={Armeni, Iro and He, Zhi-Yang and Gwak, JunYoung and Zamir, Amir R and Fischer, Martin and Malik, Jitendra and Savarese, Silvio},
  booktitle={Proceedings of the IEEE/CVF international conference on computer vision},
  pages={5664--5673},
  year={2019}
}

@inproceedings{Wald2019RIO,
    title={RIO: 3D Object Instance Re-Localization in Changing Indoor Environments},
      author    = {Johanna Wald and Armen Avetisyan and Nassir Navab and Federico Tombari and Matthias Nie{\ss}ner},
    booktitle={Proceedings IEEE International Conference on Computer Vision (ICCV)},
    year = {2019}
}

@inproceedings{3DSSG2020,
    title={Learning 3D Semantic Scene Graphs from 3D Indoor Reconstructions},
    author={Wald, Johanna and Dhamo, Helisa and Navab, Nassir and Tombari, Federico},
    booktitle={Conference on Computer Vision and Pattern Recognition (CVPR)}, 
    year={2020}
  }

@inproceedings{dai2017scannet,
    title={ScanNet: Richly-annotated 3D Reconstructions of Indoor Scenes},
    author={Dai, Angela and Chang, Angel X. and Savva, Manolis and Halber, Maciej and Funkhouser, Thomas and Nie{\ss}ner, Matthias},
    booktitle = {Proc. Computer Vision and Pattern Recognition (CVPR), IEEE},
    year = {2017}
}

@misc{agnolucci2025qualiCLIP,
      title={Quality-Aware Image-Text Alignment for Opinion-Unaware Image Quality Assessment}, 
      author={Lorenzo Agnolucci and Leonardo Galteri and Marco Bertini},
      year={2025},
      eprint={2403.11176},
      archivePrefix={arXiv},
      primaryClass={cs.CV},
      url={https://arxiv.org/abs/2403.11176}, 
}

@inproceedings{zhu20233dvista,
  title={3d-vista: Pre-trained transformer for 3d vision and text alignment},
  author={Zhu, Ziyu and Ma, Xiaojian and Chen, Yixin and Deng, Zhidong and Huang, Siyuan and Li, Qing},
  booktitle={Proceedings of the IEEE/CVF International Conference on Computer Vision},
  pages={2911--2921},
  year={2023}
}

@InProceedings{thomason2020visiondialog,
  title = 	 {Vision-and-Dialog Navigation},
  author =       {Thomason, Jesse and Murray, Michael and Cakmak, Maya and Zettlemoyer, Luke},
  booktitle = 	 {Proceedings of the Conference on Robot Learning},
  pages = 	 {394--406},
  year = 	 {2020},
  editor = 	 {Kaelbling, Leslie Pack and Kragic, Danica and Sugiura, Komei},
  volume = 	 {100},
  series = 	 {Proceedings of Machine Learning Research},
  month = 	 {30 Oct--01 Nov},
  publisher =    {PMLR},
  url = 	 {https://proceedings.mlr.press/v100/thomason20a.html}
}

@inproceedings{zhang2024dialoc,
  title={Dialoc: An iterative approach to embodied dialog localization},
  author={Zhang, Chao and Li, Mohan and Budvytis, Ignas and Liwicki, Stephan},
  booktitle={Proceedings of the IEEE/CVF Conference on Computer Vision and Pattern Recognition},
  pages={12585--12593},
  year={2024}
}

@article{chen2020scanrefer,
    title={ScanRefer: 3D Object Localization in RGB-D Scans using Natural Language},
    author={Chen, Dave Zhenyu and Chang, Angel X and Nie{\ss}ner, Matthias},
    journal={16th European Conference on Computer Vision (ECCV)},
    year={2020}
}

@inproceedings{achlioptas2020referit_3d,
    title={{ReferIt3D}: Neural Listeners for Fine-Grained 3D Object Identification in Real-World Scenes},
    author={Achlioptas, Panos and Abdelreheem, Ahmed and Xia, Fei and Elhoseiny, Mohamed and Guibas, Leonidas J.},
    booktitle={16th European Conference on Computer Vision (ECCV)},
    year={2020}
}

@inproceedings{azuma2022scanqa,
  title={ScanQA: 3D Question Answering for Spatial Scene Understanding},
  author={Azuma, Daichi and Miyanishi, Taiki and Kurita, Shuhei and Kawanabe, Motoaki},
  booktitle={Proceedings of the IEEE/CVF Conference on Computer Vision and Pattern Recognition (CVPR)},
  year={2022}
}

@article{kulesza2012determinantal,
  title={Determinantal point processes for machine learning},
  author={Kulesza, Alex and Taskar, Ben},
  journal={Foundations and Trends{\textregistered} in Machine Learning},
  volume={5},
  number={2-3},
  pages={123--286},
  year={2012},
  publisher={Emerald Publishing Limited}
}

@inproceedings{radford2021CLIP,
  title={Learning transferable visual models from natural language supervision},
  author={Radford, Alec and Kim, Jong Wook and Hallacy, Chris and Ramesh, Aditya and Goh, Gabriel and Agarwal, Sandhini and Sastry, Girish and Askell, Amanda and Mishkin, Pamela and Clark, Jack and others},
  booktitle={International conference on machine learning},
  pages={8748--8763},
  year={2021},
  organization={PmLR}
}

@article{sattler2017activesearch,
  title={Efficient \& Effective Prioritized Matching for Large-Scale Image-Based Localization},
  author={Sattler, Torsten and Leibe, Bastian and Kobbelt, Leif},
  journal={IEEE Trans. Pattern Anal. Mach. Intell.},
  volume={39},
  number={9},
  pages={1744--1756},
  year={2017}
}

@inproceedings{taira2018inloc,
  title={{InLoc}: Indoor Visual Localization with Dense Matching and View Synthesis},
  author={Taira, Hajime and Okutomi, Masatoshi and Sattler, Torsten and Cimpoi, Mircea and Pollefeys, Marc and Sivic, Josef and Pajdla, Tomas and Torii, Akihiko},
  booktitle=CVPR,
  pages={7199--7208},
  year={2018}
}

@inproceedings{sarlin2019hloc,
  title={From Coarse to Fine: Robust Hierarchical Localization at Large Scale},
  author={Sarlin, Paul-Edouard and Cadena, Cesar and Siegwart, Roland and Dymczyk, Marcin},
  booktitle=CVPR,
  pages={12716--12725},
  year={2019}
}

@inproceedings{detone2018superpoint,
  title={{SuperPoint}: Self-Supervised Interest Point Detection and Description},
  author={DeTone, Daniel and Malisiewicz, Tomasz and Rabinovich, Andrew},
  booktitle=CVPRW,
  pages={224--236},
  year={2018}
}

@inproceedings{sarlin2020superglue,
  title={{SuperGlue}: Learning Feature Matching with Graph Neural Networks},
  author={Sarlin, Paul-Edouard and DeTone, Daniel and Malisiewicz, Tomasz and Rabinovich, Andrew},
  booktitle=CVPR,
  pages={4938--4947},
  year={2020}
}

@inproceedings{kendall2015posenet,
  title={{PoseNet}: A Convolutional Network for Real-Time 6-{DOF} Camera Relocalization},
  author={Kendall, Alex and Grimes, Matthew and Cipolla, Roberto},
  booktitle=ICCV,
  pages={2938--2946},
  year={2015}
}

@inproceedings{brahmbhatt2018mapnet,
  title={Geometry-Aware Learning of Maps for Camera Localization},
  author={Brahmbhatt, Samarth and Gu, Jinwei and Kim, Kihwan and Hays, James and Kautz, Jan},
  booktitle=CVPR,
  pages={2616--2625},
  year={2018}
}

@inproceedings{brachmann2017dsac,
  title={{DSAC} -- Differentiable {RANSAC} for Camera Localization},
  author={Brachmann, Eric and Krull, Alexander and Nowozin, Sebastian and Shotton, Jamie and Michel, Frank and Gumhold, Stefan and Rother, Carsten},
  booktitle=CVPR,
  pages={6684--6692},
  year={2017}
}

@inproceedings{brachmann2018dsacpp,
  title={Learning Less is More -- {6D} Camera Localization via {3D} Surface Regression},
  author={Brachmann, Eric and Rother, Carsten},
  booktitle=CVPR,
  pages={4654--4662},
  year={2018}
}

@inproceedings{brachmann2023ace,
  title={Accelerated Coordinate Encoding: Learning to Relocalize in Minutes using {RGB} and Poses},
  author={Brachmann, Eric and Cavallari, Tommaso and Prisacariu, Victor Adrian},
  booktitle=CVPR,
  pages={5044--5053},
  year={2023}
}

@inproceedings{arandjelovic2016netvlad,
  title={{NetVLAD}: {CNN} Architecture for Weakly Supervised Place Recognition},
  author={Arandjelovic, Relja and Gronat, Petr and Torii, Akihiko and Pajdla, Tomas and Sivic, Josef},
  booktitle=CVPR,
  pages={5297--5307},
  year={2016}
}

@inproceedings{hausler2021patchnetvlad,
  title={{Patch-NetVLAD}: Multi-Scale Fusion of Locally-Global Descriptors for Place Recognition},
  author={Hausler, Stephen and Garg, Sourav and Xu, Ming and Milford, Michael and Fischer, Tobias},
  booktitle=CVPR,
  pages={14141--14152},
  year={2021}
}

@inproceedings{berton2022cosplace,
  title={Rethinking Visual Geo-Localization for Large-Scale Applications},
  author={Berton, Gabriele and Masone, Carlo and Caputo, Barbara},
  booktitle=CVPR,
  pages={4878--4888},
  year={2022}
}

@inproceedings{alibey2023mixvpr,
  title={{MixVPR}: Feature Mixing for Visual Place Recognition},
  author={Ali-bey, Amar and Chaib-draa, Brahim and Gigu{\`e}re, Philippe},
  booktitle=CVPR,
  pages={14254--14263},
  year={2023}
}

@inproceedings{sarlin2021pixloc,
  title={Back to the Feature: Learning Robust Camera Localization from Pixels to Pose},
  author={Sarlin, Paul-Edouard and Unagar, Ajaykumar and L{\"a}rsson, M{\aa}ns and Germain, Hugo and Toft, Carl and Larsson, Viktor and Pollefeys, Marc and Lepetit, Vincent and Kahl, Lars and Sattler, Torsten},
  booktitle=CVPR,
  pages={3247--3257},
  year={2021}
}

@inproceedings{sarlin2023orienternet,
  title={{OrienterNet}: Visual Localization in {2D} Public Maps with Neural Matching},
  author={Sarlin, Paul-Edouard and DeTone, Daniel and Yang, Tsun-Yi and Avetisyan, Armen and Straub, Julian and Malisiewicz, Tomasz and Rota Bulo, Samuel and Newcombe, Richard and Kontschieder, Peter and Balntas, Vasileios},
  booktitle=CVPR,
  pages={21632--21642},
  year={2023}
}

@inproceedings{kolmet2022text2pos,
  title={{Text2Pos}: Text-to-Point-Cloud Cross-Modal Localization},
  author={Kolmet, Manuel and Zhou, Qunjie and Osep, Aljosa and Leal-Taix{\'e}, Laura},
  booktitle=CVPR,
  pages={15172--15181},
  year={2022}
}

@inproceedings{anderson2018r2r,
  title={Vision-and-Language Navigation: Interpreting Visually-Grounded Navigation Instructions in Real Environments},
  author={Anderson, Peter and Wu, Qi and Teney, Damien and Bruce, Jake and Johnson, Mark and S{\"u}nderhauf, Niko and Reid, Ian and Gould, Stephen and van den Hengel, Anton},
  booktitle=CVPR,
  pages={3674--3683},
  year={2018}
}

@inproceedings{qi2020reverie,
  title={{REVERIE}: Remote Embodied Visual Referring Expression in Real Indoor Environments},
  author={Qi, Yuankai and Wu, Qi and Anderson, Peter and Wang, Xin and Wang, William Yang and Shen, Chunhua and van den Hengel, Anton},
  booktitle=CVPR,
  pages={9982--9991},
  year={2020}
}

@inproceedings{hong2021vlnbert,
  title={{VLN-BERT}: A Recurrent Vision-and-Language {BERT} for Navigation},
  author={Hong, Yicong and Wu, Qi and Qi, Yuankai and Rodriguez-Opazo, Cristian and Gould, Stephen},
  booktitle=CVPR,
  pages={1643--1653},
  year={2021}
}

@inproceedings{zhou2024navgpt,
  title={{NavGPT}: Explicit Reasoning in Vision-and-Language Navigation with Large Language Models},
  author={Zhou, Gengze and Hong, Yicong and Wu, Qi},
  booktitle=AAAI,
  pages={7641--7649},
  year={2024}
}

@inproceedings{wu2021scenegraphfusion,
  title={{SceneGraphFusion}: Incremental {3D} Scene Graph Prediction from {RGB-D} Sequences},
  author={Wu, Shun-Cheng and Wald, Johanna and Tateno, Keisuke and Navab, Nassir and Tombari, Federico},
  booktitle=CVPR,
  pages={7515--7525},
  year={2021}
}

@inproceedings{gu2024conceptgraphs,
  title={{ConceptGraphs}: Open-Vocabulary {3D} Scene Graphs for Perception and Planning},
  author={Gu, Qiao and Kuwajerwala, Alihusein and Morin, Sacha and Jatavallabhula, Krishna Murthy and Sen, Bipasha and Agarwal, Aditya and Rivera, Corban and Paul, William and Ellis, Kirsty and Chellappa, Rama and Gan, Chuang and de Melo, Celso and Tenenbaum, Joshua B. and Torralba, Antonio and Shkurti, Florian and Paull, Liam},
  booktitle={IEEE Int. Conf. Robot. Automat. (ICRA)},
  year={2024}
}

@inproceedings{peng2023openscene,
  title={{OpenScene}: {3D} Scene Understanding with Open Vocabularies},
  author={Peng, Songyou and Genova, Kyle and Jiang, Chiyu Max and Tagliasacchi, Andrea and Pollefeys, Marc and Funkhouser, Thomas},
  booktitle=CVPR,
  pages={815--824},
  year={2023}
}

@inproceedings{kerr2023lerf,
  title={{LERF}: Language Embedded Radiance Fields},
  author={Kerr, Justin and Kim, Chung Min and Goldberg, Ken and Kanazawa, Angjoo and Tancik, Matthew},
  booktitle=ICCV,
  pages={19729--19739},
  year={2023}
}

@inproceedings{ma2023sqa3d,
  title={{SQA3D}: Situated Question Answering in {3D} Scenes},
  author={Ma, Xiaojian and Yong, Silong and Zheng, Zilong and Li, Qing and Liang, Yitao and Zhu, Song-Chun and Huang, Siyuan},
  booktitle=ICLR,
  year={2023}
}

@inproceedings{hong2023_3dllm,
  title={{3D-LLM}: Injecting the {3D} World into Large Language Models},
  author={Hong, Yining and Zhen, Haoyu and Chen, Peihao and Zheng, Shuhong and Du, Yilun and Chen, Zhenfang and Gan, Chuang},
  booktitle=NeurIPS,
  year={2023}
}

@article{qwen2vl,
  title={Qwen2-{VL}: Enhancing Vision-Language Model's Perception of the World at Any Resolution},
  author={Wang, Peng and Bai, Shuai and Tan, Sinan and Wang, Shijie and Fan, Zhihao and Bai, Jinze and Chen, Keqin and Liu, Xuejing and Wang, Jialin and Ge, Wenbin and others},
  journal={arXiv preprint arXiv:2409.12191},
  year={2024}
}

@article{mikolov2013word2vec,
  title={Efficient estimation of word representations in vector space},
  author={Mikolov, Tomas and Chen, Kai and Corrado, Greg and Dean, Jeffrey},
  journal={arXiv preprint arXiv:1301.3781},
  year={2013}
}
\end{document}